\documentclass[runningheads]{llncs}
\usepackage{graphicx}
\usepackage{comment}
\usepackage{amsmath,amssymb} 
\usepackage{color}
\usepackage{subfigure}
\usepackage{mmstyles}
\usepackage{array}
\usepackage{multirow}

\makeatletter
\def\hlinew#1{%
  \noalign{\ifnum0=`}\fi\hrule \@height #1 \futurelet
   \reserved@a\@xhline}
\makeatother

\newcommand{\tabincell}[2]{\begin{tabular}{@{}#1@{}}#2\end{tabular}}

\newcolumntype{P}[1]{>{\raggedright\arraybackslash}p{#1}}
\newcolumntype{M}[1]{>{\centering\arraybackslash}m{#1}}

\begin{document}
\pagestyle{headings}
\mainmatter
\def\ECCVSubNumber{1486}  

\title{Thinking in Frequency: Face Forgery Detection by Mining Frequency-aware Clues} 


\titlerunning{F$^3$-Net: Frequency in Face Forgery Network}
\author{Yuyang Qian$^{\ast, \dagger}$ \inst{1,2} \and
Guojun Yin$^{\dagger, \ddagger}$ \inst{1} \and
Lu Sheng$^{\ddagger}$ \inst{3} \and
Zixuan Chen$^{\ast}$ \inst{1,4} \and
Jing Shao\inst{1}}
\authorrunning{Y. Qian et al.}
%
\institute{
SenseTime Research, \and
University of Electronic Science and Technology of China, \and
College of Software, Beihang University, \and
Northwestern Polytechnical University\\
\email{qyy@std.uestc.edu.cn, zixuan.sean.chen@hotmail.com, lsheng@buaa.edu.cn, \{yinguojun, shaojing\}@sensetime.com, }
}

\renewcommand{\thefootnote}{\fnsymbol{footnote}}
\footnotetext[1]{This work was done during the internship of Yuyang Qian and Zixuan Chen at SenseTime Research.}
\footnotetext[4]{The first two authors contributed equally.}
\footnotetext[5]{Corresponding Author.}

\maketitle


\begin{abstract}

As realistic facial manipulation technologies have achieved remarkable progress, social concerns about potential malicious abuse of these technologies bring out an emerging research topic of face forgery detection. However, it is extremely challenging since recent advances are able to forge faces beyond the perception ability of human eyes, especially in compressed images and videos. We find that mining forgery patterns with the awareness of frequency could be a cure, as frequency provides a complementary viewpoint where either subtle forgery artifacts or compression errors could be well described. To introduce frequency into the face forgery detection, we propose a novel Frequency in Face Forgery Network (F$^3$-Net), taking advantages of two different but complementary frequency-aware clues, 1) frequency-aware decomposed image components, and 2) local frequency statistics, to deeply mine the forgery patterns via our two-stream collaborative learning framework. We apply DCT as the applied frequency-domain transformation. Through comprehensive studies, we show that the proposed F$^3$-Net significantly outperforms competing state-of-the-art methods on all compression qualities in the challenging FaceForensics++ dataset, especially wins a big lead upon low-quality media.

\end{abstract}


\section{Introduction}
\label{sec:intro}

Rapid development of deep learning driven generative models~\cite{gansurvey,ganlarge,ganprogressive,ganstyle,ganstar} enables an attacker to create, manipulate or even forge the media of a human face (\ie, images and videos, etc.) that cannot be distinguished even by human eyes.
However, malicious distribution of forged media would cause security issues and even crisis of confidence in our society.
Therefore, it is supremely important to develop effective face forgery detection methods.

Various methods~\cite{mesonet,fdftnet,multitask,capsule,social,CNNfingerprints,twostream,CNNwild} have been proposed to detect the forged media. 
A series of earlier works relied on hand-crafted features~\eg, local pattern analysis~\cite{patternanalysis}, noise variances evaluation~\cite{noise} and steganalysis features~\cite{stegfeature,richmodel} to discover forgery patterns and magnify faint discrepancy between real and forged images.
Deep learning introduces another pathway to tackle this challenge, recent learning-based forgery detection methods~\cite{ff++17,xception} tried to mine the forgery patterns in feature space using convolutional neural networks (CNNs), having achieved remarkable progresses on public datasets, \eg, FaceForensics++~\cite{faceforensics++}.

\begin{figure}[tp]
\centering
\includegraphics[width=\linewidth]{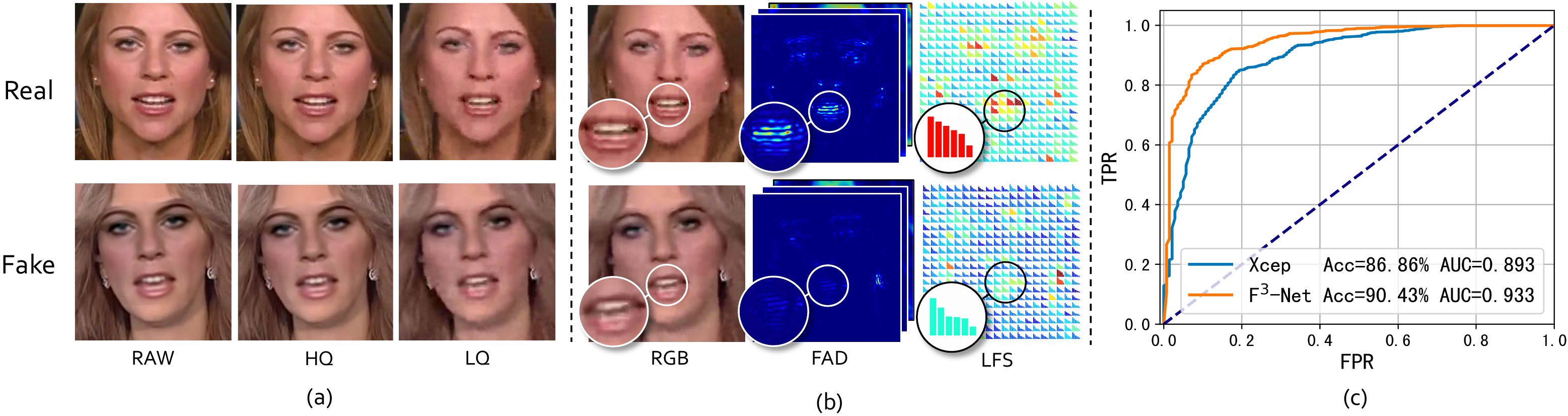}
\caption{Frequency-aware tampered clues for face forgery detection. (a) RAW, high quality (HQ) and low quality (LQ) real and fake images with the same identity, manipulation artifacts are barely visible in low quality images. (b) Frequency-aware forgery clues in low quality images using the proposed \emph{Frequency-aware Decomposition (FAD)} and \emph{Local Frequency Statistics (LFS)}. (c) ROC Curve of the proposed \textbf{Frequency in Face Forgery Network (F$^3$-Net)} and baseline (\ie, Xception~\cite{xception}). The proposed F$^3$-Net wins the Xception with a large margin. Best viewed in color.}
\label{fig: Advantages}
\end{figure}

Current state-of-the-art face manipulation algorithms, such as DeepFake~\cite{gitdeepfake}, FaceSwap~\cite{gitfaceswap}, Face2Face~\cite{face2face} and NeuralTextures~\cite{nt}, have been able to conceal the forgery artifacts, so that it becomes extremely difficult to discover the flaws of these refined counterfeits, as shown in Fig.~\ref{fig: Advantages}(a).
What's worse, if the visual quality of a forged face is tremendously degraded, such as compressed by JPEG or H.264 with a large compression ratio, the forgery artifacts will be contaminated by compression error, and sometimes cannot be captured in RGB domain any more.
Fortunately, these artifacts can be captured in frequency domain, as many prior studies suggested~\cite{CNNfingerprints,liveFreq,deepfrequent,simpleFrequencyFeatrues,easytospot}, in the form of unusual frequency distributions when compared with real faces.
However, how to involve frequency-aware clues into the deeply learned CNN models? This question also raises alongside.
Conventional frequency domains, such as FFT and DCT, do not match the shift-invariance and local consistency owned by nature images, thus vanilla CNN structures might be infeasible.
As a result, CNN-compatible frequency representation becomes pivotal if we would like to leverage the discriminative representation power of learnable CNNs for frequency-aware face forgery detection.
To this end, we would like to introduce two frequency-aware forgery clues that are compatible with the knowledge mining by deep convolutional networks.

From one aspect, it is possible to decompose an image by separating its frequency signals, while each decomposed image component indicates a certain band of frequencies.
The first frequency-aware forgery clue is thus discovered by the intuition that we are able to identify subtle forgery artifacts that are somewhat salient (\ie, in the form of unusual patterns) in the decomposed components with higher frequencies, as the examples shown in the middle column of Fig.~\ref{fig: Advantages}(b).
This clue is compatible with CNN structures, and is surprisingly robust to compression artifacts.
From the other aspect, the decomposed image components describe the frequency-aware patterns in the spatial domain, but not explicitly render the frequency information directly in the neural networks.
We suggest the second frequency-aware forgery clue as the local frequency statistics.
In each densely but regularly sampled local spatial patch, the statistics is gathered by counting the mean frequency responses at each frequency band.
These frequency statistics re-assemble back to a multi-channel spatial map, where the number of channels is identical to the number of frequency bands.
As shown in the last column of Fig.~\ref{fig: Advantages}(b), the forgery faces have distinct local frequency statistics than the corresponding real ones, even though they look almost the same in the RGB images.
Moreover, the local frequency statistics also follows the spatial layouts as the input RGB images, thus also enjoy effective representation learning powered by CNNs.
Meanwhile, since the decomposed image components and local frequency statistics are complementary to each other but both of them share inherently similar frequency-aware semantics, thus they can be progressively fused during the feature learning process.

Therefore, we propose a novel \textbf{Frequency in Face Forgery Network (F$^3$-Net)}, that capitalizes on the aforementioned frequency-aware forgery clues.
The proposed framework is composed of two frequency-aware branches, one aims at learning subtle forgery patterns through \emph{Frequency-aware Image Decomposition (FAD)}, and the other would like to extract high-level semantics from \emph{Local Frequency Statistics (LFS)} to describe the frequency-aware statistical discrepancy between real and forged faces.
These two branches, are further gradually fused through a cross-attention module, namely \emph{MixBlock}, which encourages rich interactions between the aforementioned FAD and LFS branches.
The whole face forgery detection model is learned by the cross-entropy loss in an end-to-end manner.
Extensive experiments demonstrate that the proposed F$^3$-Net significantly improves the performance over low-quality forgery media with a thorough ablation study.
We also show that our framework largely exceeds competing state-of-the-arts on all compression qualities in the challenging FaceForensics++~\cite{faceforensics++}. As shown in Fig.\ref{fig: Advantages}(c), the effectiveness and superiority of the proposed frequency-aware F$^{3}$-Net is obviously demonstrated by comparing the ROC curve with Xception~\cite{xception}(baseline, previous state-of-the-art seeing in Sec.\ref{sec:exp}). 
Our contributions in this paper are summarized as follows:

\noindent \textbf{1) Frequency-aware Decomposition (FAD)} aims at learning frequency-aware forgery patterns through frequency-aware image decomposition. The proposed FAD module adaptively partitions the input image in the frequency domain according to learnable frequency bands and represents the image with a series of frequency-aware components.

\noindent \textbf{2) Local Frequency Statistics (LFS)} extracts local frequency statistics to describe the statistical discrepancy between real and fake faces.
The localized frequency statistics not only reveal the unusual statistics of the forgery images at each frequency band, but also share the structure of natural images, and thus enable effective mining through CNNs.

\noindent \textbf{3)} The proposed framework collaboratively learns the frequency-aware clues from FAD and LFS, by a cross-attention (a.k.a MixBlock) powered two-stream networks.
The proposed method achieves the state-of-the-art performance on the challenging FaceForensics++ dataset~\cite{faceforensics++}, especially wins a big lead in the low quality forgery detection.


\section{Related Work}
\label{sec:relatedwork}

With the development of computer graphics and neural networks especially generative adversarial networks (GANs)~\cite{gansurvey,ganlarge,ganprogressive,ganstyle,ganstar}, 
face forgery detection has gained more and more interest in our society. 
%
%
Various attempts have been made for face forgery detection and achieved remarkable progress, but learning-based generation methods such as NeuralTextures~\cite{nt} are still difficult to detect because they introduce only small-scale subtle visual artifacts especially in low quality videos. To address the problem, various additional information is used to enhance performance.

\vspace{0.1cm}
\noindent \textbf{Spatial-Based Forgery Detection.}
To address face forgery detection tasks, a variety of methods have been proposed. 
Most of them are based on the spatial domain such as RGB and HSV. Some approaches~\cite{excolorcues,ff++23} exploit specific artifacts arising from the synthesis process such as color or shape cues.
Some studies~\cite{colorcomponents,colorcues,ela} extract color-space features to classify fake and real images. For example, ELA~\cite{ela} uses pixel-level errors to detect image forgery.
%
%
Early methods~\cite{ff++10,ff++17} use hand-crafted features for shallow CNN architectures. 
Recent methods~\cite{mesonet,fdftnet,multitask,capsule,social,CNNfingerprints,twostream,CNNwild} use deep neural networks to extract high-level information from the spatial domain and get remarkable progress. 
MesoInception-4~\cite{mesonet} is a CNN-based Network inspired by InceptionNet~\cite{inception} to detect forged videos. GANs Fingerprints Analysis~\cite{CNNfingerprints} introduces deep manipulation discriminator to discover specific manipulation patters. However, most of them use only spatial domain information and therefore are not sensitive to subtle manipulation clues that are difficult to detect in color-space. In our works, we take advantage of frequency cues to mine small-scale detailed artifacts that are helpful especially in low-quality videos.

%

\vspace{0.1cm}
\noindent \textbf{Frequency-Based Forgery Detection.} Frequency domain analysis is a classical and important method in image signal processing and has been widely used in a number of applications such as image classification~\cite{frequencylayer,wavefeature,DCT}, 
steganalysis~\cite{fdsteganalysis,phaseaware}, texture classification~\cite{rotation,wavelet,gabortexture} and super-resolution~\cite{fastcnn,waveletsrnet}. 
Recently, several attempts have been made to solve forgery detection using frequency cues. Some studies use Wavelet Transform (WT)~\cite{wavelettransform} or Discrete Fourier Transform (DFT)~\cite{detectgan,easytospot,simpleFrequencyFeatrues} to convert pictures to frequency domain and mine underlying artifacts.
%
For example, Durall \etal~\cite{simpleFrequencyFeatrues} extracts frequency-domain information using DFT transform and averaging the amplitudes of different frequency bands. 
Stuchi \etal~\cite{frequencylayer} uses a set of fixed frequency domain filters to extract different range of information followed by a fully connected layer to get the output.
Besides, filtering, a classic image signal processing method, is used to refine and mine underlying subtle information in forgery detection, which leverages existing knowledge of the characteristics of fake images.
Some studies use high-pass filters~\cite{easytospot,autoencoder,videoforgery,noisesurvey}, Gabor filters~\cite{phaseaware,gabortexture} etc. to extract features of interest (\eg edge and texture information) based on features regarding with high frequency components.
Phase Aware CNN~\cite{phaseaware} uses hand-crafted Gabor and high-pass filters to augment the edge and texture features. 
Universal Detector~\cite{easytospot} finds that significant differences can be obtained in the spectrum between real and fake images after high-pass filtering.
However, the filters used in these studies are often fixed and hand-crafted thus fail to capture the forgery patterns adaptively. In our work, we make use of frequency-aware image decomposition to mine frequency forgery cues adaptively.






\section{Our Approach}
\label{sec:method}

\begin{figure}[tp]
\centering
\includegraphics[width=\linewidth]{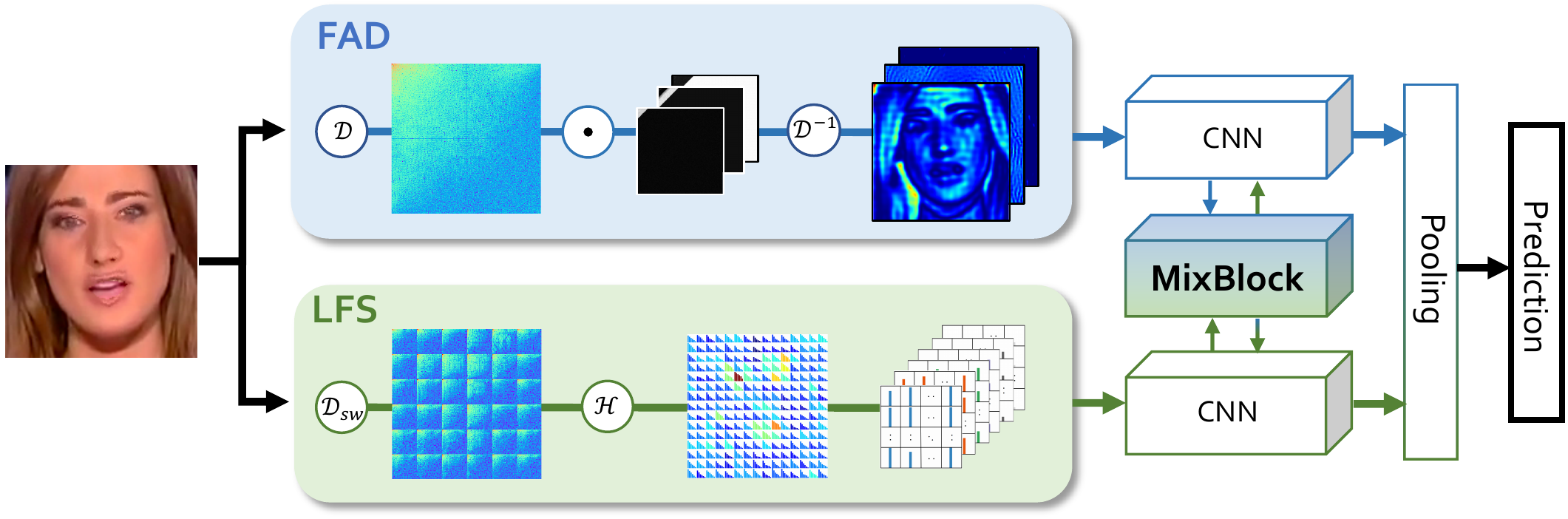}
\caption{Overview of the F$^3$-Net. The proposed architecture consists of three novel methods: \emph{FAD} for learning subtle manipulation patterns through frequency-aware image decomposition; \emph{LFS} for extracting local frequency statistics and \emph{MixBlock} for collaborative feature interaction.}
\label{fig: pipeline}
\end{figure}

In this section, we introduce the proposed two kinds of frequency-aware forgery clue mining methods, \ie, frequency-aware decomposition (in Sec.~\ref{subsec:FFP}) and local frequency statistics (in Sec.~\ref{subsec:lfh}), and then present the proposed cross-attention two-stream collaborative learning framework (in Sec.~\ref{subsec:mixblock}).

\subsection{FAD: Frequency-Aware Decomposition}
\label{subsec:FFP}


Towards the frequency-aware image decomposition, former studies usually apply hand-crafted filter banks~\cite{phaseaware,gabortexture} in the spatial domain, thus fail to cover the complete frequency domain.
Meanwhile, the fixed filtering configurations make it hard to adaptively capture the forgery patterns.
To this end, we propose a novel frequency-aware decomposition (FAD), to adaptively partition the input image in the frequency domain according to a set of learnable frequency filters.
The decomposed frequency components can be inversely transformed to the spatial domain, resulting in a series of frequency-aware image components.
These components are stacked along the channel axis, and then inputted into a convolutional neural network (in our implementation, we employ an Xception~\cite{xception} as the backbone) to comprehensively mine forgery patterns.

To be specific, we manually design $N$ binary base filters $\{\mathbf{f}_{base}^i\}_{i=1}^N$ (or called masks) that explicitly partition the frequency domain into low, middle and high frequency bands.
And then we add three learnable filters $\{\mathbf{f}^i_w\}_{i=1}^N$ to these base filters.
The frequency filtering is a dot-product between the frequency response of the input image and the combined filters $\mathbf{f}_{base}^i + \sigma(\mathbf{f}_w^i), i=\{1, \ldots, N\}$,
where $\sigma(x) = \frac{1-\exp(-x)}{1+\exp(-x)}$ aims at squeezing $x$ within the range between $-1$ and $+1$.
Therefore, to an input image $\mathbf{x}$, the decomposed image components are obtained by

\begin{figure}[tp]
\centering
\includegraphics[width=\linewidth]{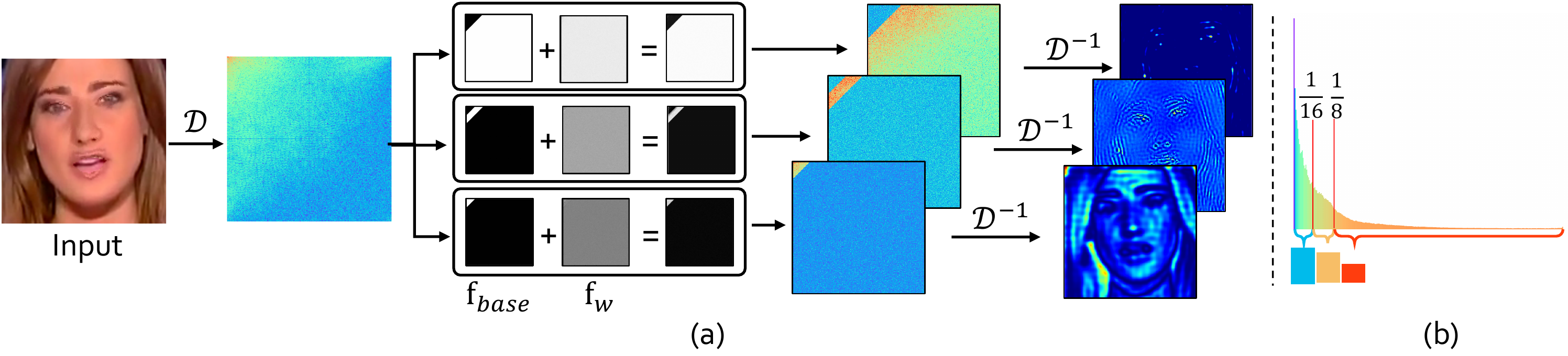}
\caption{(a) The proposed \emph{Frequency-aware Decomposition (FAD)} to discover salient frequency components. $\mathcal{D}$ indicates applying Discrete Cosine Transform (DCT). $\mathcal{D}^{-1}$ indicates applying Inversed Discrete Cosine Transform (IDCT). Several frequency band components can be concatenated together to extract a wider range of information. (b) The distribution of the DCT power spectrum. We flatten 2D power spectrum to 1D by summing up the amplitudes of each frequency band. We divide the spectrum into 3 bands with roughly equal energy.}
\label{fig: FFP}
\end{figure}

\begin{equation}
	\mathbf{y}_i = \mathcal{D}^{-1} \{ \mathcal{D}(\mathbf{x}) \odot [\mathbf{f}_{base}^i + \sigma(\mathbf{f}_w^i)] \}, ~~i = \{1, \ldots, N\}.
\end{equation}
$\odot$ is the element-wise product.
We apply $\mathcal{D}$ as the Discrete Cosine Transform (DCT)~\cite{dcttransform}, according to its wide applications in image processing, and its nice layout of the frequency distribution, \ie, low-frequency responses are placed in the top-left corner, and high-frequency responses are located in the bottom-right corner.
Moreover, recent compression algorithms, such as JPEG and H.264, usually apply DCT in their frameworks, thus DCT-based FAD will be more compatible towards the description of compression artifacts out of the forgery patterns.
Observing the DCT power spectrum of natural images, we find that the spectral distribution is non-uniform and most of the amplitudes are concentrated in the low frequency area.
We apply the base filters $\mathbf{f}_{base}$ to divide the spectrum into $N$ bands with roughly equal energy, from low frequency to high frequency.
The added learnable $\{\mathbf{f}_w^i\}_{i=1}^N$ provides more adaptation to select the frequency of interest beyond the fixed base filters.
Empirically, as shown in Fig.~\ref{fig: FFP}(b), the number of bands $N=3$,
%
%
the low frequency band $\mathbf{f}_{base}^1$ is the first $1/16$ of the entire spectrum, 
the middle frequency band $\mathbf{f}_{base}^2$ is between $1/16$ and $1/8$ of the spectrum, 
and the high frequency band $\mathbf{f}_{base}^3$ is the last $7/8$.

\subsection{LFS: Local Frequency Statistics}
\label{subsec:lfh}

The aforementioned FAD has provided frequency-aware representation that is compatible with CNNs, but it has to represent frequency-aware clues back into the spatial domain, thus fail to directly utilize the frequency information.
Also knowing that it is usually infeasible to mine forgery artifacts by extracting CNN features directly from the spectral representation, we then suggest to estimate local frequency statistics (LFS) to not only explicitly render frequency statistics but also match the shift-invariance and local consistency that owned by natural RGB images.
These features are then inputted into a convolutional neural network, \ie, Xception~\cite{xception}, to discover high-level forgery patterns.

\begin{figure}[tp]
\centering
\includegraphics[width=0.99\linewidth]{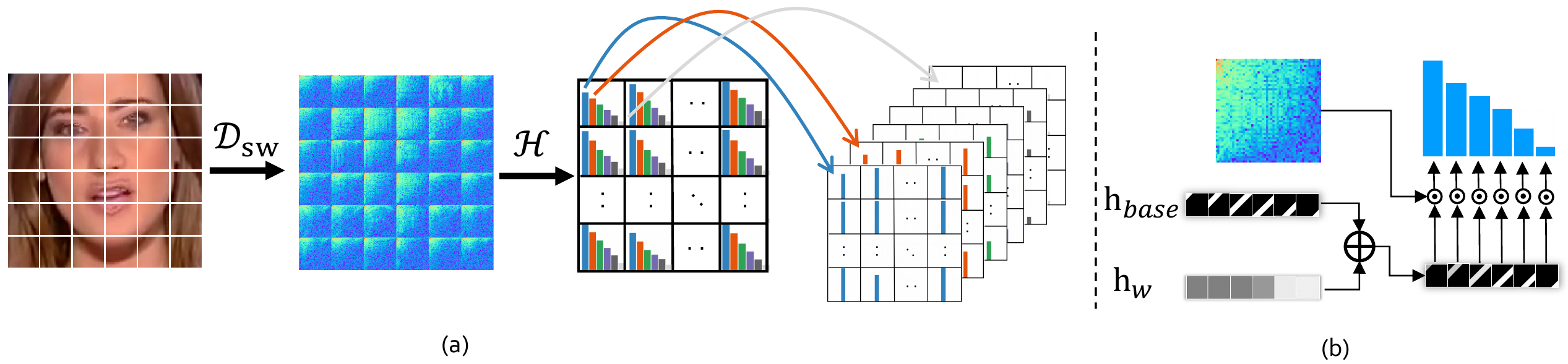}
\caption{(a) The proposed \emph{Local Frequency Statistics (LFS)} to extract local frequency domain statistical information. SWDCT indicates applying Sliding Window Discrete Cosine Transform and $\mathcal{H}$ indicates gathering statistics on each grid adaptively. (b) Extracting statistics from a DCT power spectrum graph, $\oplus$ indicates element-wise addition and $\odot$ indicates element-wise multiplication.}
\label{fig: LFH}
\end{figure}

As shown in Fig.~\ref{fig: LFH}(a), we first apply a Sliding Window DCT (SWDCT) on the input RGB image (\ie, taking DCTs densely on sliding windows of the image) to extract the localized frequency responses, and then counting the mean frequency responses at a series of learnable frequency bands.
These frequency statistics re-assemble back to a multi-channel spatial map that shares the same layout as the input image.
This LFS provides a localized aperture to detect detailed abnormal frequency distributions.
Calculating statistics within a set of frequency bands allows a reduced statistical representation, whilst yields a smoother distribution without the interference of outliers.

To be specific, in each window $\mathbf{p} \in \mathbf{x}$ , after DCT, the local statistics is gathered in each frequency band, which is constructed similarly as the way used in FAD (see Sec.~\ref{subsec:FFP}).
In each band, the statistics become 
\begin{equation}
\mathbf{q}_i = \log_{10} \| \mathcal{D}(\mathbf{p}) \odot [\mathbf{h}_{base}^i + \sigma(\mathbf{h}_w^i)] \|_1, ~~i = \{1,\ldots, M\},
\end{equation}
Note that $\log_{10}$ is applied to balance the magnitude in each frequency band.
The frequency bands are collected by equally partitioning the spectrum in to M parts, following the order from low frequency to high frequency.
Similarly, $\mathbf{h}_{base}^i$ is the base filter, $\mathbf{h}_w^i$ is the learnable filter, $i = \{ 1,\ldots,M\}$.
The local frequency statistics $\mathbf{q}$ for a window $\mathbf{p}$ is then transposed as a $1\times1\times M$ vector.
These statistics vectors gathered from all windows are re-assembled into a matrix with downsampled spatial size of the input image, whose number of channels is equal to M.
This matrix will act as the input to the later convolutional layers.

Practically in our experiments, we empirically adopt the window size as $10$, the sliding stride as $2$, and the number of bands as $M=6$, thus the size of the output matrix will be $149\times149\times6$ if the input image is of size $299\times299\times3$.

\begin{figure}[tp]
\centering
\includegraphics[width=\linewidth]{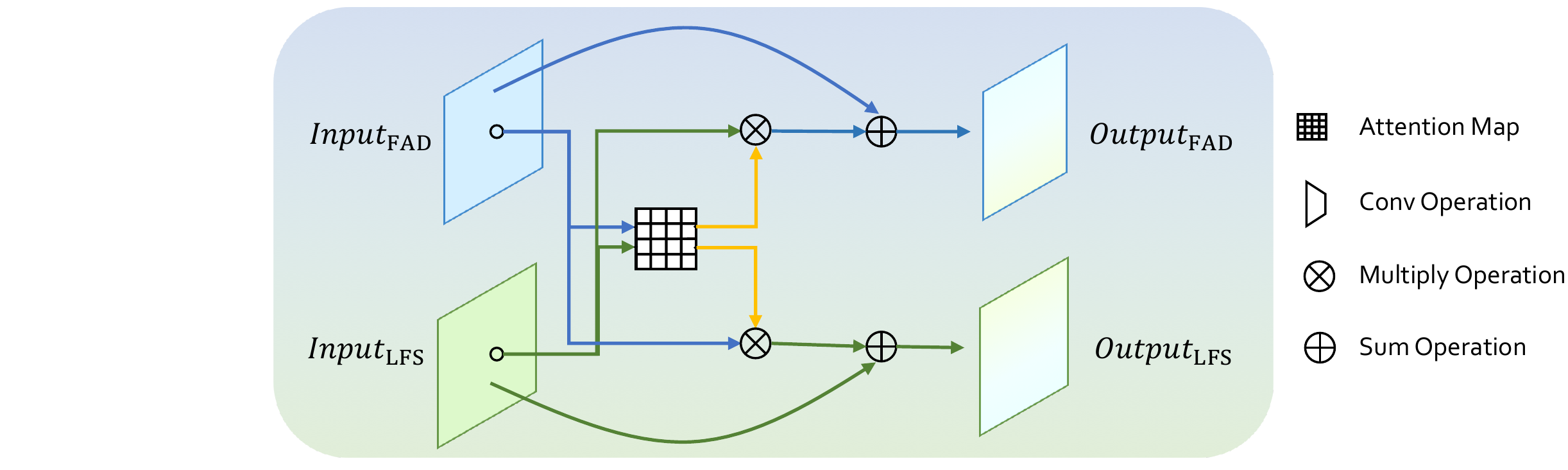}
\caption{The proposed \emph{MixBlock}. $\otimes$ indicates matrix multiplication and $\oplus$ indicates element-wise addition.}
\label{fig: mixblock}
\end{figure}

\subsection{Two-stream Collaborative Learning Framework}
\label{subsec:mixblock}

As mentioned in Sec.~\ref{subsec:FFP} and Sec.~\ref{subsec:lfh}, the proposed FAD and LFS modules mine the frequency-aware forgery clues from two different but inherently connected aspects.
We argue that these two types of clues are different but complementary.
Thus, we propose a collaborative learning framework that powered by cross-attention modules, to gradually fuse two-stream FAD and LFS features.
To be specific, the whole network architecture of our F$^3$-Net is composed of two branches equipped with Xception blocks~\cite{xception}, one is for the decomposed image components generated by FAD, and the other is for local frequency statistics generated by LFS, as shown in Fig.~\ref{fig: pipeline}.

We propose a cross-attention fusion module for the feature interaction and message passing every several Xception blocks. 
As shown in Fig.~\ref{fig: mixblock}, different from the simple concatenation widely used in previous methods~\cite{slowfast,deepfrequent,twostream}, we firstly calculate the cross-attention weight using the feature maps from the two branches.
The cross-attention matrix is adopted to augment the attentive features from one stream to another.

In our implementation, we use Xception network~\cite{xception} pretrained on the ImageNet~\cite{imagenet} for both branches, each of which has 12 blocks.
The newly-introduced layers and blocks are randomly initialized. 
The cropped face is adopted as the input of the framework after resized as $299 \times 299$.
Empirically, we adopt MixBlock after block 7 and block 12 to fuse two types of frequency-aware clues according to their mid-level and high-level semantics.
We train the F$^3$-Net by the well-known cross entropy loss, and the whole system can be trained in an end-to-end fashion.


\section{Experiment}
\label{sec:exp}

\begin{table}[tp]
    \footnotesize
    \centering
    \caption{Quantitative results on FaceForensics++ dataset with all quality settings, \ie LQ indicates low quality (heavy compression), HQ indicates high quality (light compression) and RAW indicates raw videos without compression. The bold results are the best. Note that Xception+ELA and Xception-PAFilters are two Xception baselines that are equipped with ELA~\cite{ela} and PAFilters~\cite{phaseaware}.}
    \label{table:comparisonwithprevous}
    \begin{tabular}{P{3.6cm} M{1.3cm} M{1.3cm} M{1.3cm} M{1.3cm} M{1.3cm} M{1.3cm}}
    \hlinew{1.1pt}
    Methods &\tabincell{c}{Acc\\(LQ)} & \tabincell{c}{AUC\\(LQ)} & \tabincell{c}{Acc\\(HQ)} & \tabincell{c}{AUC\\(HQ)} & \tabincell{c}{Acc\\(RAW)} & \tabincell{c}{AUC\\(RAW)} \\ 
    \hline
    Steg.Features~\cite{richmodel}& 55.98\% & - & 70.97\% & -  & 97.63\% & - \\ 
    LD-CNN~\cite{ff++17}& 58.69\% & - & 78.45\% & -  & 98.57\% & -  \\
    Constrained Conv~\cite{ff++10}& 66.84\% & -   & 82.97\% & -& 98.74\% & - \\
    CustomPooling CNN~\cite{ff++51}& 61.18\% & -  & 79.08\% & -& 97.03\% & -  \\
    MesoNet~\cite{mesonet} & 70.47\% & -  & 83.10\% & - & 95.23\% & -\\ 
    Face X-ray~\cite{xray}  & - & 0.616  & - & 0.874& - & - \\ 
    Xception~\cite{xception}& 86.86\% & 0.893  & 95.73\% & 0.963& 99.26\% & 0.992  \\
    Xception-ELA~\cite{ela} & 79.63\% & 0.829 & 93.86\% & 0.948 & 98.57\% & 0.984 \\
    Xception-PAFilters~\cite{phaseaware} & 87.16\% & 0.902 & - & - & - & - \\
    F$^{3}$-Net (Xception)  & \textbf{90.43\%} & \textbf{0.933} & \textbf{97.52\%} & \textbf{0.981}  & \textbf{99.95\%} & \textbf{0.998}\\
    \hline
    \hline
    Optical Flow~\cite{flo} & 81.60\% & - & - & - & - & - \\ 
    Slowfast~\cite{slowfast} & 90.53\% & 0.936  & 97.09\% & 0.982 & 99.53\% & 0.994 \\ 
    
    F$^{3}$-Net(Slowfast) & \textbf{93.02\%} & \textbf{0.958} & \textbf{98.95\%} & \textbf{0.993}  & \textbf{99.99\%} & \textbf{0.999} \\ 
    \hlinew{1.1pt}
    \end{tabular}
    \end{table}

\subsection{Setting}

\noindent\textbf{Dataset.}
Following previous face forgery detection methods~\cite{CNNRNN,simpleFrequencyFeatrues,xray,flo}, we conduct our experiments on the challenging FaceForensics++~\cite{faceforensics++} dataset.
FaceForensics++ is a face forgery detection video dataset containing 1,000 real videos, in which 720 videos are used for training, 140 videos are reserved for validation and 140 videos for testing.
Most videos contain frontal faces without occlusions and were collected from YouTube with the consent of the subjects.
Each video undergoes four manipulation methods to generate four fake videos, therefore there are 5,000 videos in total.
The number of frames in each video is between 300 and 700.
The size of the real videos is augmented four times to solve category imbalance between the real and fake data.
270 frames are sampled from each video, following the setting as in FF++~\cite{faceforensics++}. 
Output videos are generated with different quality levels, so as to create a realistic setting for manipulated videos, \ie, RAW, High Quality (HQ) and Low Quality (LQ), respectively.
%

We use the face tracking method proposed by Face2Face~\cite{face2face} to crop the face and adopt a conservative crop to enlarge the face region by a factor of $1.3$ around the center of the tracked face, following the setting in~\cite{faceforensics++}.

\vspace{0.1cm}
\noindent \textbf{Evaluation Metrics.}
We apply the Accuracy score (Acc) and Area Under the Receiver Operating Characteristic Curve (AUC) as our evaluation metrics.
(1) \textbf{Acc}. Following FF++~\cite{faceforensics++}, we use the accuracy score as the major evaluation metric in our experiments.
This metric is commonly used in face forgery detection tasks~\cite{mesonet,stegfeature,capsule}.
Specifically, for single-frame methods, we average the accuracy scores of each frame in a video. 
%
%
%
%
(2) \textbf{AUC}.
Following face X-ray~\cite{xray}, we use AUC score as another evaluation metric.
For single-frame methods, we also average the AUC scores of each frame in a video.

\vspace{0.1cm}
\noindent \textbf{Implementation Details.}
In our experiments, we use Xception~\cite{xception} pretrained on the ImageNet~\cite{imagenet} as backbone for the proposed F$^{3}$-Net. The newly-introduced layers and blocks are randomly initialized. 
%
The networks are optimized via SGD. 
We set the base learning rate as 0.002 and use Cosine~\cite{cosine} learning rate scheduler.
The momentum is set as 0.9. The batch size is set as 128. We train for about $150k$ iterations.

Some studies~\cite{CNNRNN,flo} use videos as the input of the face forgery detection system.
To demonstrate the generalization of the proposed methods, we also plug LFS and FAD into existing video-based methods, \ie Slowfast-R101~\cite{slowfast} pre-trained on Kinetics-400~\cite{kinetics}.
The networks are optimized via SGD.
We set the base learning rate as 0.002. The momentum is set as 0.9. The batch size is set as 64. We train the model for about $200k$ iterations.

\subsection{Comparing with previous methods}

In this section, on the FaceForensics++ dataset, we compare our method with previous face forgery detection methods.
\begin{table}[tp]
\footnotesize
\centering
\caption{Quantitative results (Acc) on FaceForensics++ (LQ) dataset with four manipulation methods, \ie DeepFakes(DF)~\cite{gitdeepfake}, Face2Face(F2F)~\cite{face2face}, FaceSwap(FS)~\cite{gitfaceswap} and NeuralTextures(NT)~\cite{nt}). The bold results are the best.}
\label{table:comparisonsingle}
\begin{tabular}{P{3.6cm} M{1.5cm} M{1.5cm} M{1.5cm} M{1.5cm}}
\hlinew{1.1pt}
Methods & DF~\cite{gitdeepfake} & F2F~\cite{face2face} & FS~\cite{gitfaceswap} & NT~\cite{nt} \\ 
\hline
Steg.Features~\cite{richmodel} & 67.00\%  & 48.00\%  & 49.00\% & 56.00\%  \\ 
LD-CNN~\cite{ff++17} & 75.00\% & 56.00\%  & 51.00\% & 62.00\%   \\
Constrained Conv~\cite{ff++10}& 87.00\% & 82.00\% & 74.00\% & 74.00\% \\
CustomPooling CNN~\cite{ff++51}& 80.00\% & 62.00\% & 59.00\% & 59.00\% \\
MesoNet~\cite{mesonet} & 90.00\% & 83.00\% & 83.00\% & 75.00\% \\
Xception~\cite{xception}& 96.01\% & 93.29\% & 94.71\% & 79.14\% \\ 
F$^{3}$-Net(Xception) & \textbf{97.97\%} & \textbf{95.32\%} & \textbf{96.53\%} & \textbf{83.32\%} \\ 
\hline \hline
Slowfast~\cite{slowfast} & 97.53\% & 94.93\% & 95.01\% & 82.55\% \\ 
F$^{3}$-Net(Slowfast) & \textbf{98.62\%} & \textbf{95.84\%} & \textbf{97.23\%} & \textbf{86.01\%} \\
\hlinew{1.1pt}
\end{tabular}
\end{table}

\vspace{+1mm}
\noindent\textbf{Evaluations on Different Quality Settings.}
The results are listed in Tab.\ref{table:comparisonwithprevous}.
The proposed F$^3$-Net outperforms all the reference methods on all quality settings, \ie, LQ, HQ and RAW, respectively.
According to the low-quality (LQ) setting, the proposed F$^{3}$-Net achieves $90.43\%$ in Acc and $0.933$ in AUC respectively, with a remarkable improvement comparing to the current state-of-the-art methods, \ie, about 3.5\% performance gain on Acc score against the best performed reference method (\ie, Xception-PAFilters with 87.16\% \emph{v.s.} F$^3$-Net with 90.43\%). 
The performance gains mainly benefit from the information mining from frequency-aware FAD and LFS clues, which helps the proposed F$^{3}$-Net more capable of detecting subtle manipulation artifacts as well as robust to heavy compression errors than plain RGB-based networks.
It is worth noting that some methods~\cite{ela,simpleFrequencyFeatrues,multitask,phaseaware} also try to employ complementary information from other domains, and try to take advantages of prior knowledge.
For example, Steg.Features~\cite{richmodel} employs hand-crafted steganalysis features and PAFilters~\cite{phaseaware} tries to augment the edge and texture features by hand-crafted Gabor and high-pass filters. 
Different from these methods, the proposed F$^{3}$-Net makes good use of CNN-friendly and adaptive mechanism to augment the FAD and LFS module, thus significantly boost the performance by a considerable margin

\begin{figure}[t]
\centering
\subfigure[]{
\begin{minipage}{0.48\linewidth}
\centering
\includegraphics[width=0.99\linewidth]{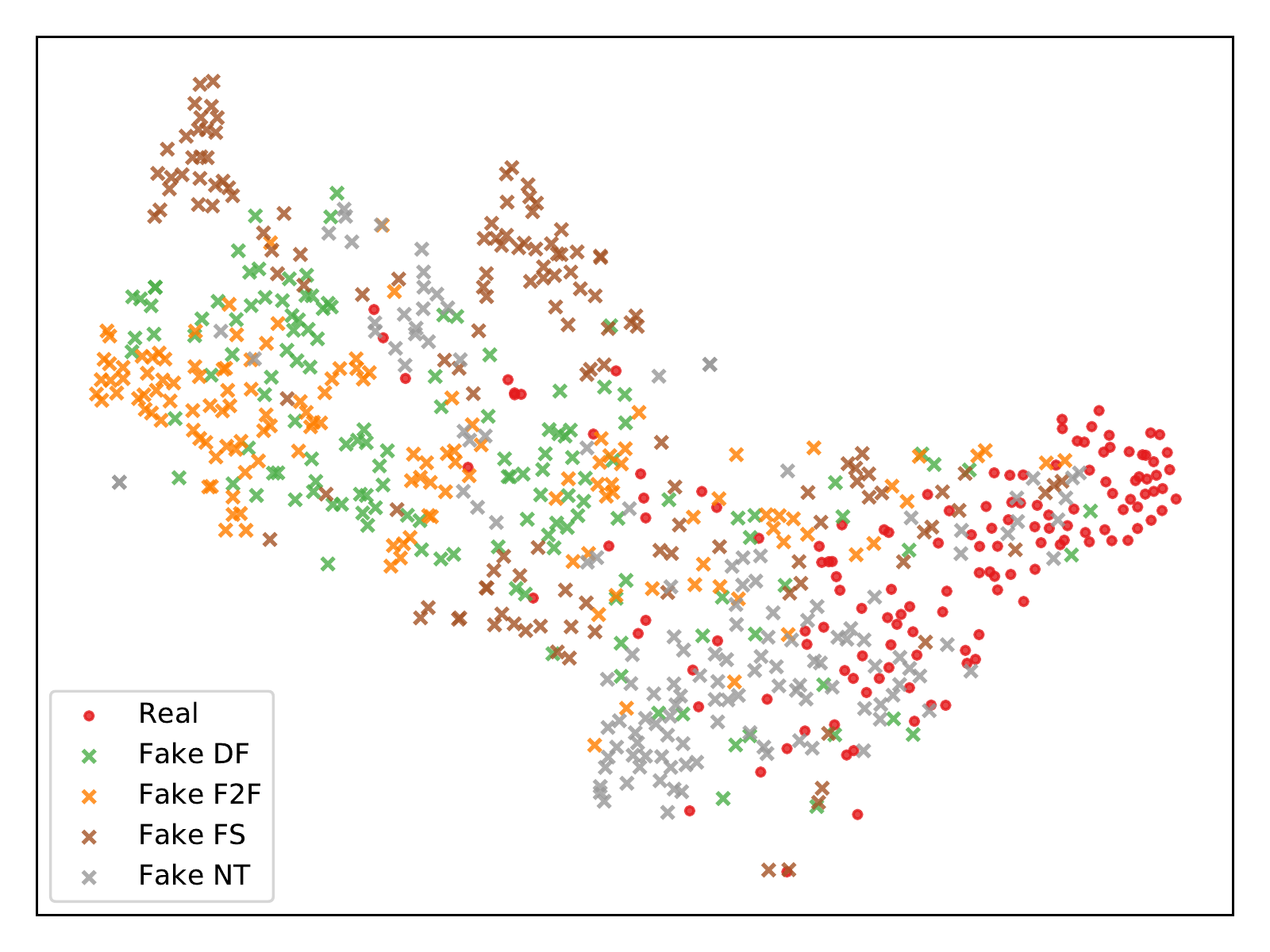}
\end{minipage}}
\subfigure[]{
        \begin{minipage}{0.48\linewidth}
\centering
\includegraphics[width=0.99\linewidth]{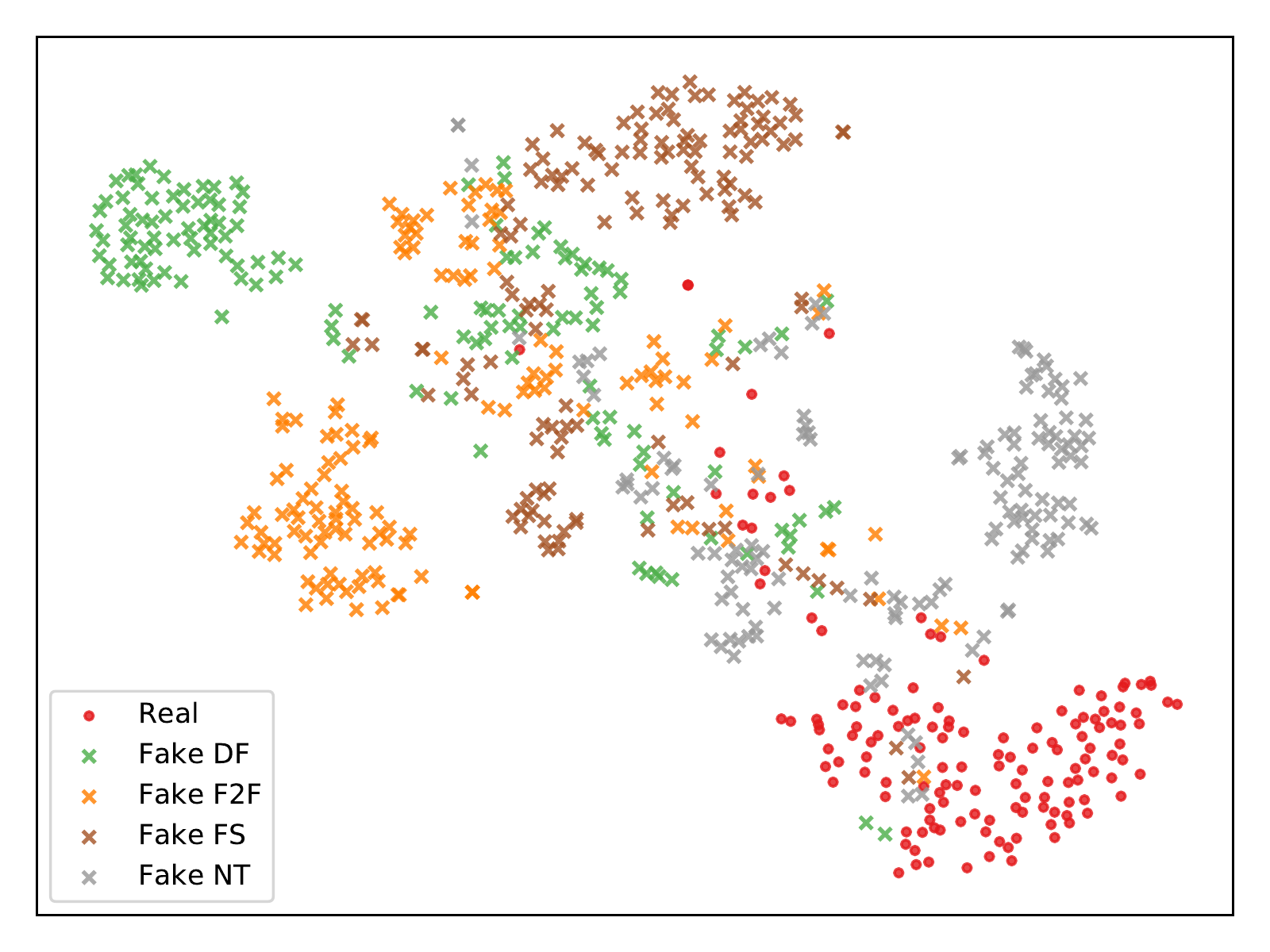}
\end{minipage}}
\caption{The t-SNE embedding visualization of the baseline (a) and F$^{3}$-Net (b) on FaceForensics++~\cite{faceforensics++} low quality (LQ) task. Red color indicates the real videos, the rest colors represent data generated by different manipulation methods. Best viewed in color.}
\label{fig: tsne}
\end{figure}

\vspace{+1mm}
\noindent\textbf{Towards Different Manipulation Types.}
Furthermore, we evaluate the proposed F$^{3}$-Net on different face manipulation methods listed in~\cite{faceforensics++}.
The models are trained and tested exactly on the low quality videos from one face manipulation methods.
The results are shown in Tab.~\ref{table:comparisonsingle}.
%
Of the four manipulation methods, the videos generated by NeuralTextures (NT)~\cite{nt} is extremely challenging due to its excellent generation performance in synthesizing realistic faces without noticeable forgery artifacts. 
The performance of our proposed method is particularly impressive when detecting forged faces by NT, leading to an improvement of about $4.2\%$ on the Acc score, against the baseline method Xception~\cite{xception}.

Furthermore, we also showed the t-SNE~\cite{tsne} feature spaces of data in FaceForensics++~\cite{faceforensics++} low quality (LQ) task, by the Xception and our F$^3$-Net, as shown in Fig.~\ref{fig: tsne}. 
Xception cannot divide the real data and NT-based forged data since their features are cluttered in the t-SNE embedding space, as shown in Fig.~\ref{fig: tsne}(a).
However, although the feature distances between real videos and NT-based forged videos are closer than the rest pairs in the feature space of F$^3$-Net, they are still much farther away than those in the feature space of Xception.
It, from another viewpoint, proves that the proposed F$^3$-Net can mine effective clues to distinguish the real and forged media.

\vspace{+1mm}
\noindent\textbf{Video-based Extensions.}
Meanwhile, there are also several studies~\cite{CNNRNN,flo} using multiple frames as the input.
To evaluate the generalizability of our methods, we involve the proposed LFS and FAD into Slowfast-R101~\cite{slowfast} due to its excellent performance for video classification.
The results are shown in Tab.\ref{table:comparisonwithprevous} and Tab.\ref{table:comparisonsingle}.
More impressively, our F$^{3}$-Net (Slowfast) achieves the better performances than the baseline using Slowfast only, \ie, $93.02\%$ and $0.958$ of Acc and AUC scores in comparison to $90.53\%$ and $0.936$, in low quality (LQ) task, as shown in Tab.~\ref{table:comparisonwithprevous}.
Slowfast-F$^{3}$-Net also wins over 3\% on the NT-based manipulation, as shown in Tab.~\ref{table:comparisonsingle}, not to mention the rest three manipulation types.
These excellent performances further demonstrate the effectiveness of our proposed frequency-aware face forgery detection method.

\begin{figure}[t]
\begin{minipage}[b]{0.5\linewidth}
\centering
\subtable[]{
\begin{tabular}{c|ccc|cc}
\hlinew{1.1pt}
ID & FAD & LFS & MixBlock & Acc & AUC \\ 
\hline
1 & - & - & - & 86.86\% & 0.893 \\  
2 & $\surd$ & - & - & 87.95\% & 0.907 \\ 
3 & - & $\surd$ & - & 88.73\% & 0.920 \\
4 & $\surd$ & $\surd$ & - & 89.89\% & 0.928 \\ 
5 & $\surd$ & $\surd$ & $\surd$ & \textbf{90.43\%} & \textbf{0.933} \\
\hlinew{1.1pt}
\end{tabular}}
\end{minipage}
\begin{minipage}[hb]{0.5\linewidth}
\centering
\subfigure[]{
\includegraphics[width=\linewidth]{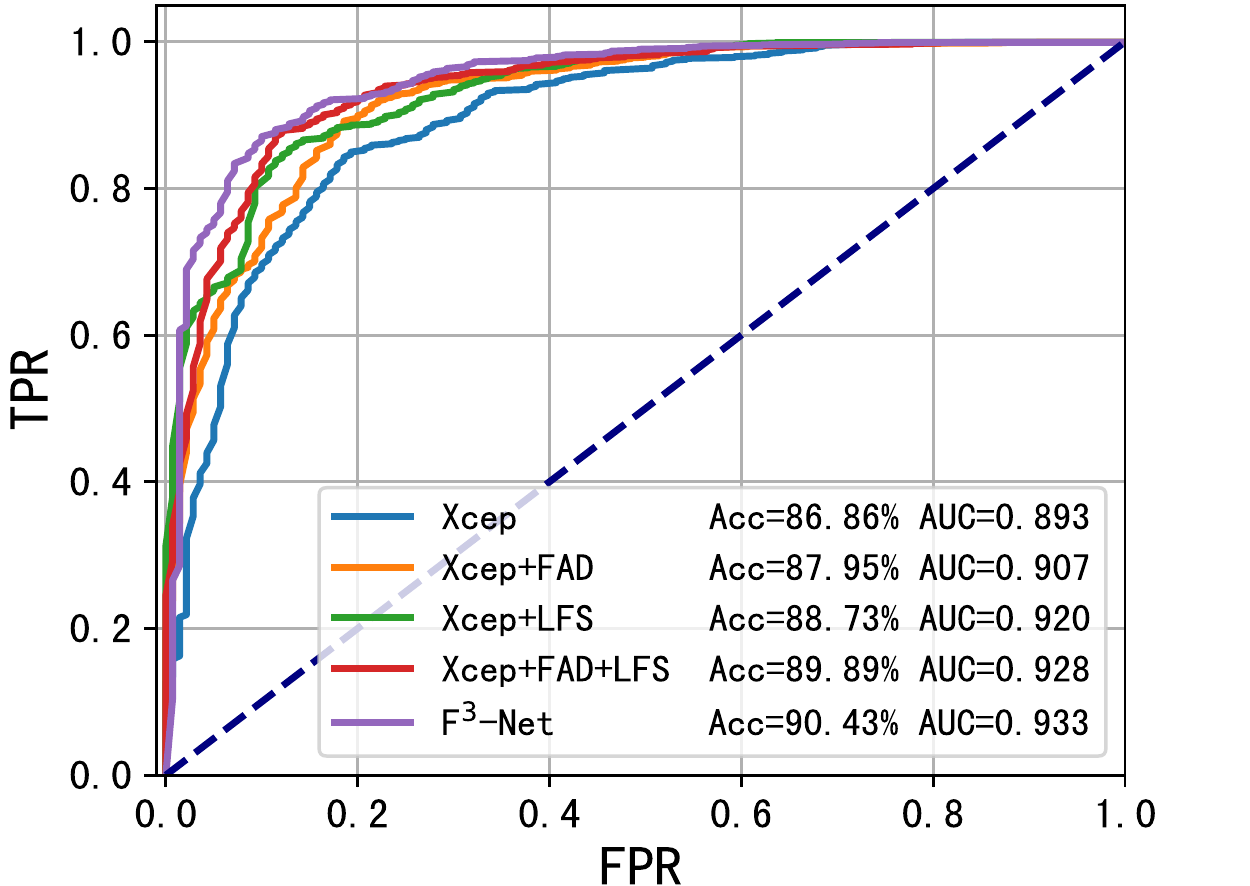}
}
\end{minipage}
\caption{(a) Ablation study of the proposed F$^{3}$-Net on the low quality task(LQ). We compare F$^{3}$-Net and its variants by removing the proposed FAD, LFS and MixBlock step by step. (b) ROC Curve of the models in our ablation studies.}
\label{tab: roc}
\end{figure}

\subsection{Ablation Study}

\begin{figure}[tp]
    \centering
    \includegraphics[width=\linewidth]{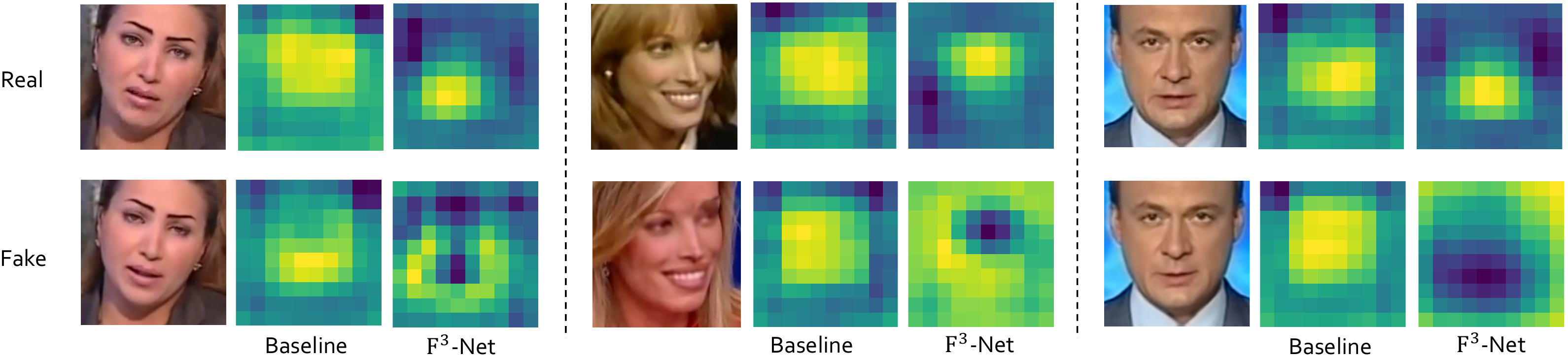}
    \caption{The visualization of the feature map extracted by baseline (\ie, Xception) and the proposed F$^{3}$-Net respectively.}
    \label{fig: featmap}
\end{figure}

\noindent \textbf{Effectiveness of LFS, FAD and MixBlock.}
To evaluate the effectiveness of the proposed LFS, FAD and MixBlock, we quantitatively evaluate F$^{3}$-Net and its variants: 1) the baseline (Xception), 2) F$^{3}$-Net w/o LFS and MixBlock, 3) F$^{3}$-Net w/o FAD and MixBlock, 4) F$ ^{3}$-Net w/o MixBlock. 

The quantitative results are listed in Fig.~\ref{tab: roc}(a). By comparing model 1 (baseline) and model 2 (Xception with FAD), the proposed FAD consistently improves the Acc and AUC scores.
When adding the LFS (model 4) based on model 2, the Acc and AUC scores become even higher.
Plugging MixBlock (model 5) into the two branch structure (model 4) gets the best performance, $90.43\%$ and $0.933$ for Acc and AUC scores, respectively.
These progressively improved performances validate that the proposed FAD and LFS module indeed helps the forgery detection, and they are complementary to each other.
MixBlock introduce more advanced cooperation between FAD and LFS, and thus introduce additional gains.
As shown in the ROC curves in Fig.~\ref{tab: roc}(b), F$^{3}$-Net receives the best performance at lower false positive rate (FPR), while low FPR rate is a most challenging scenario to forgery detection system.
%
To better understand the effectiveness of the proposed methods, we visualize the feature maps extracted by the baseline (Xception) and $F^{3}$-Net, respectively, as shown in Fig.~\ref{fig: featmap}. 
%
The discriminativeness of these feature maps is obviously improved by the proposed F$^{3}$-Net, \eg, there are clear differences between real and forged faces in the feature distributions of F$^{3}$-Net, while the corresponding feature maps generated by Xception are similar and indistinguishable.

\vspace{0.1cm}
\noindent \textbf{Ablation study on FAD.}
%
To demonstrate the benefits of adaptive frequency decomposition on complete frequency domain in FAD, 
we evaluate the proposed FAD and its variants by removing or replacing some components, \ie, 1) Xception (baseline), 2) a group of hand-drafted filters used in Phase Aware CNN~\cite{phaseaware}, denoted as Xception + PAFilters, 3) proposed FAD without learnable filters, denoted as Xception + FAD ($\mathbf{f}_{base}$), and 4) Xception with the full FAD, denoted as Xception+FAD ($\mathbf{f}_{base}+\mathbf{f}_w$).
All the experiments are under the same hyper-parameters for fair comparisons. 
As shown in the left part of Tab.~\ref{table:fixFAD}, the performance of Xception is improved by a considerable margin on the Acc and AUC scores after applying the FAD, in comparison with other methods using fixed filters (Xception + PAFilters).
If the learnable filters are removed, there will also be a sudden performance drop.

\begin{table}[tp]
\footnotesize
\centering
\caption{Ablation study and component analysis on FAD in FF++ low quality (LQ) tasks. Left: comparing traditional fixed filters with the proposed FAD. Right: comparing FAD and its variants with different kinds of frequency components. We use the full FAD in model FAD-All.
}
\label{table:fixFAD}
\begin{tabular}{P{4.0cm} M{1.2cm} M{1.4cm}||P{1.6cm} M{1.2cm} M{1.3cm}}
\hlinew{1.1pt}
Models & Acc & AUC & Models & Acc & AUC \\ 
\hline
 Xception & 86.86\% & 0.893 & FAD-Low & 86.95\% & 0.901\\
 Xception+PAFilters~\cite{phaseaware} & 87.16\% & 0.902 & FAD-Mid & 87.57\% & 0.904 \\ 
 Xception+FAD ($\mathbf{f}_{base}$) & 87.12\% & 0.901 & FAD-High & 87.77\% & 0.906  \\
 Xception+FAD ($\mathbf{f}_{base}+\mathbf{f}_w$) & 87.95\% & 0.907 & FAD-All & 87.95\% & 0.907 \\
\hlinew{1.1pt}
\end{tabular}
\end{table}

We further demonstrate the importance of extracting complete information from complete frequency domain by quantitatively evaluating FAD with different kinds of frequency components, \ie, 1) FAD-Low, FAD with low frequency band components, 2) FAD-Mid, FAD with middle frequency band components, 3) FAD-High, FAD with high frequency band components and 4) FAD-All, FAD with all frequency bands components.
The quantitative results are listed in the right part of Tab.~\ref{table:fixFAD}.
By comparing FAD-Low, FAD-Mid and FAD-High, the model with high frequency band components achieves the best scores, which indicates that high frequency clues are indubitably helpful for forgery detection.
It is because high-frequent clues are usually correlated with forgery-sensitive edges and textures.
After making use of all three kinds of information (\ie, FAD-All), we achieved the highest result.
Since the low frequency components preserve the global picture, the middle and high frequency reveals the small-scale detailed information, concatenating them together helps to obtain richer frequency-aware clues and is able to mine forgery patterns more comprehensively.

\begin{table}[tp]
\centering
\caption{Ablation study on LFS. Here we use only LFS branch and add components step by step. SWDCT indicates using Sliding Window DCT instead of traditional DCT, Stat indicates adopting frequency statistics and D-Stat indicates using our proposed adaptive frequency statistics.}
\label{table: lflhcmp}
\begin{tabular}{M{1.3cm}| M{1.3cm} | M{1.3cm} | M{1.15cm} M{1.15cm} M{1.15cm} M{1.15cm}M{1.15cm}M{1.15cm}}
\hlinew{1.1pt}
SWDCT & Stat & D-Stat &\tabincell{c}{Acc\\(LQ)} & \tabincell{c}{AUC\\(LQ)} & \tabincell{c}{Acc\\(HQ)} & \tabincell{c}{AUC\\(HQ)} & \tabincell{c}{Acc\\(RAW)} & \tabincell{c}{AUC\\(RAW)} \\ 
\hline
- & - & - & 76.16\% & 0.724 & 90.12\% & 0.905 & 95.28\% & 0.948 \\
$\surd$ & - & -  & 82.47\% & 0.838 & 93.85\% & 0.940 & 97.02\% & 0.964  \\
$\surd$ & $\surd$ & - & 84.89\% & 0.865 & 94.12\% & 0.936 & 97.97\% & 0.975  \\
$\surd$ & $\surd$ & $\surd$ & 86.16\% & 0.889 & 94.76\% & 0.951 & 98.37\% & 0.983 \\
\hlinew{1.1pt}
\end{tabular}
\end{table}

\vspace{0.1cm}
\noindent \textbf{Ablation study on LFS.}
%
To demonstrate the effectiveness of SWDCT and dynamic statistical strategy in the proposed LFS introduced in Sec.~\ref{subsec:lfh}, 
we take the experiments (Xception as backbone) on the proposed LFS and its variants, 1) Baseline, of which the frequency spectrum of the full image by traditional DCT; 2) SWDCT, adopting the localized frequency response by SWDCT ; 3) SWDCT+Stat, adopting the general statistical strategy with filters $\mathbf{h}_{base}$; 4) SWDCT+Stat+D-Stat, the proposed FAD consisted of SWDCT and the adaptive frequency statistics with learnable filters $\mathbf{h}_w$.
The results are shown in Tab.~\ref{table: lflhcmp}. 
Comparing with traditional DCT operation on the full image, the proposed SWDCT significantly improves the performance by a large margin since it is more sensitive to the spatial distributions of the local statistics, and letting the Xception back capture the forgery clues.
The improvement of using the statistics is significant and local statistics are more robust to unstable or noisy spectra, especially when optimized by adding the adaptive frequency statistics.


\section{Conclusions}

In this paper, we propose an innovative face forgery detection framework that can make use of frequency-aware forgery clues, named as F$^{3}$-Net.
The proposed framework is composed of two frequency-aware branches, one focuses on mining subtle forgery patterns through frequency components partition, and the other aims at extracting small-scale discrepancy of frequency statistics between real and forged images. Meanwhile, a novel cross-attention module is applied for two-stream collaborative learning.
Extensive experiments demonstrate the effectiveness and significance of the proposed F$^{3}$-Net on FaceForencis++ dataset, especially in the challenging low quality task.


~\\


\noindent \textbf{Acknowledgements.} \ This work is supported by SenseTime Group Limited, in part by key research and development program of Guangdong Province, China, under grant 2019B010154003. The corresponding authors are Guojun Yin and Lu Sheng. The contribution of Yuyang Qian and Guojun Yin are Equal.

%
%
\bibliographystyle{splncs04}
\bibliography{1486}

\clearpage


\section{Appendix}
\label{sec:supple}

\vspace{0.1cm}
\subsection{Greedy-searched Threshold for Accuracy}
The accuracy  value is influenced by the threshold $\theta$.  If the output score $s$ of the model is above $\theta$, the instance is recognized as \texttt{fake}, otherwise, the video is classified as \texttt{real}. In most previous works on the binary classification tasks, the parameter $\theta$ is set as $0.5$ by default.  
In our work, we find that the parameter $\theta$ makes a big difference on the exact value of the metric Acc. 
Therefore, we greedily search the highest accuracy value in the validation set of FF++ dataset with different thresholds in the parameter set $\{0,0.01,0.02,0.03,...,0.98,0.99,1\}$ for the best threshold $\theta_{max}$ and then evaluate on the test set with the greedy-searched threshold $\theta_{max}$ . 

In Tab.\ref{table:comparisonwithprevous} and Tab.\ref{table:comparisonsingle} in the main paper, the Acc values of our method and the previous methods we re-implemented, \ie Xception~\cite{xception}, Xception-ELA~\cite{ela}, Xception-PAFilters~\cite{phaseaware}, F$^{3}$-Net (Xception), Slowfast~\cite{slowfast} and F$^{3}$-Net(Slowfast) are calculated by the greedy-searched thresholds  $\theta_{max}$ respectively.
The comparasion of accuracy values with the greedy-searched threshold  $\theta_{max}$ and standard threshold  $\theta_{0.5}$ are shown in Tab.\ref{table:comparisonwithprevous_acc_thres} and Tab.\ref{table:accurancy_threshold_thres}. 
Compared with the standard threshold $\theta=0.5$, the accuracy values with greedy-searched threshold $\theta_{max}$ are higher under all of the settings, especially when the classifier performs not very well. Regardless of the greedy-serarched threshold, comparing our method F$^3$-Net and previous methods in Tab.\ref{table:comparisonwithprevous_0.5} and Tab.\ref{table:comparisonsingle_0.5}, it is also demonstrated the priority and effectiveness of our proposed F$^3$-Net using the standard threshold $0.5$.

\begin{table}
\scriptsize
\centering
\caption{Quantitative results (Acc) with  greedy-searched threshold and standard threshold on FaceForensics++ dataset with all quality settings, \ie LQ indicates low quality (heavy compression), HQ indicates high quality (light compression) and RAW indicates raw videos without compression. The bold results are the best. Note that Xception+ELA and Xception-PAFilters are two Xception baselines that are equipped with ELA~\cite{ela} and PAFilters~\cite{phaseaware}. The values in the brackets are the accuracy values with $\theta=0.5$ and the value drops compared with the greedy-searched threshold.}
\label{table:comparisonwithprevous_acc_thres}
\begin{tabular}{P{2.9cm} M{3.1cm} M{3.1cm} M{3.1cm}}
\hlinew{1.1pt}
Methods &\tabincell{c}{Acc(LQ)} &  \tabincell{c}{Acc(HQ)} & \tabincell{c}{Acc(RAW)}\\ 
\hline
Xception~\cite{xception}& 86.86\%(82.71\%,4.15\%$\downarrow$) & 95.73\%(95.04\%,0.69\%$\downarrow$) & 99.26\%(98.77\%,0.49\%$\downarrow$)  \\
Xception-ELA~\cite{ela} & 79.63\%(73.69\%,5.94\%$\downarrow$) & 93.86\%(92.09\%,1.77\%$\downarrow$) & 98.57\%(97.13\%,1.44\%$\downarrow$)  \\
Xception-PAFilters~\cite{phaseaware} & 87.16\%(83.24\%,3.92\%$\downarrow$) &  - & - \\
F$^{3}$-Net (Xception)  & \textbf{90.43\%}(86.89\%,3.54\%$\downarrow$) & \textbf{97.52\%}(97.31\%,0.21\%$\downarrow$)  & \textbf{99.95\%}(99.84\%,0.11\%$\downarrow$) \\
\hline
\hline
Slowfast~\cite{slowfast} & 90.53\%(88.25\%,2.28\%$\downarrow$) & 97.09\%(96.92\%,0.17\%$\downarrow$) & 99.53\%(99.34\%,0.19\%$\downarrow$)  \\ 
F$^{3}$-Net(Slowfast) & \textbf{93.02\%}(92.37\%,0.65\%$\downarrow$) & \textbf{98.95\%}(98.64\%,0.31\%$\downarrow$) & \textbf{99.99\%}(99.91\%,0.08\%$\downarrow$) \\ 
\hlinew{1.1pt}
\end{tabular}
\end{table}
\begin{table}
\scriptsize
\centering
\caption{Quantitative results (Acc) with  greedy-searched threshold and standard threshold on FaceForensics++ (LQ) dataset with four manipulation methods, \ie DeepFakes(DF)~\cite{gitdeepfake}, Face2Face(F2F)~\cite{face2face}, FaceSwap(FS)~\cite{gitfaceswap} and NeuralTextures(NT)~\cite{nt}). The bold results are the best. The values in the brackets are the accuracy values with $\theta=0.5$ .}
\label{table:accurancy_threshold_thres}
\begin{tabular}{P{2.3cm} M{2.3cm} M{2.3cm} M{2.3cm} M{2.3cm}}
\hlinew{1.1pt}
Methods & DF~\cite{gitdeepfake} & F2F~\cite{face2face} & FS~\cite{gitfaceswap} & NT~\cite{nt} \\ 
\hline
Xception~\cite{xception}& 96.01\% (94.27\%) & 93.29\% (91.98\%) & 94.71\% (93.03\%) & 79.14\% (76.43\%) \\ 
F$^{3}$-Net(Xception) & \textbf{97.97\%} (96.81\%) & \textbf{95.32\%} (94.01\%) & \textbf{96.53\%} (95.85\%) & \textbf{83.32\%} (79.36\%) \\ 
\hline \hline
Slowfast~\cite{slowfast} & 97.53\% (96.01\%) & 94.93\% (92.47\%) & 95.01\% (93.86\%) & 82.55\% (80.07\%) \\ 
F$^{3}$-Net(Slowfast) & \textbf{98.62\%} (98.54\%) & \textbf{95.84\%} (93.91\%) & \textbf{97.23\%} (96.82\%) & \textbf{86.01\%} (83.74\%) \\
\hlinew{1.1pt}
\end{tabular}
\end{table}
\begin{table}
\footnotesize
\centering
\caption{Quantitative results (Acc with $\theta=0.5$ ) on FaceForensics++ dataset with all quality settings, \ie LQ indicates low quality (heavy compression), HQ indicates high quality (light compression) and RAW indicates raw videos without compression. The bold results are the best. Note that Xception+ELA and Xception-PAFilters are two Xception baselines that are equipped with ELA~\cite{ela} and PAFilters~\cite{phaseaware}.}
\label{table:comparisonwithprevous_0.5}
\begin{tabular}{P{3.6cm} M{1.3cm} M{1.3cm} M{1.3cm} M{1.3cm} M{1.3cm} M{1.3cm}}
\hlinew{1.1pt}
Methods &\tabincell{c}{Acc\\(LQ)} & \tabincell{c}{AUC\\(LQ)} & \tabincell{c}{Acc\\(HQ)} & \tabincell{c}{AUC\\(HQ)} & \tabincell{c}{Acc\\(RAW)} & \tabincell{c}{AUC\\(RAW)} \\ 
\hline
Steg.Features~\cite{richmodel}& 55.98\% & - & 70.97\% & -  & 97.63\% & - \\ 
LD-CNN~\cite{ff++17}& 58.69\% & - & 78.45\% & -  & 98.57\% & -  \\
Constrained Conv~\cite{ff++10}& 66.84\% & -   & 82.97\% & -& 98.74\% & - \\
CustomPooling CNN~\cite{ff++51}& 61.18\% & -  & 79.08\% & -& 97.03\% & -  \\
MesoNet~\cite{mesonet} & 70.47\% & -  & 83.10\% & - & 95.23\% & -\\ 
Face X-ray~\cite{xray}  & - & 0.616  & - & 0.874& - & - \\ 
Xception~\cite{xception}& 82.71\% & 0.893  & 95.04\% & 0.963& 98.77\% & 0.992  \\
Xception-ELA~\cite{ela} & 73.69.\% & 0.829 & 92.09\% & 0.948 & 97.13\% & 0.984 \\
Xception-PAFilters~\cite{phaseaware} & 83.24\% & 0.902 & - & - & - & - \\
F$^{3}$-Net (Xception)  & \textbf{86.89\%} & \textbf{0.933} & \textbf{97.31\%} & \textbf{0.981}  & \textbf{99.84\%} & \textbf{0.998}\\
\hline
\hline
Optical Flow~\cite{flo} & 81.60\% & - & - & - & - & - \\ 
Slowfast~\cite{slowfast} & 88.25\% & 0.936  & 96.92\% & 0.982 & 99.34\% & 0.994 \\ 

F$^{3}$-Net(Slowfast) & \textbf{92.37\%} & \textbf{0.958} & \textbf{98.64\%} & \textbf{0.993}  & \textbf{99.91\%} & \textbf{0.999} \\ 
\hlinew{1.1pt}
\end{tabular}
\end{table}
\begin{table}
\footnotesize
\centering
\caption{Quantitative results (Acc with $\theta=0.5$) on FaceForensics++ (LQ) dataset with four manipulation methods, \ie DeepFakes(DF)~\cite{gitdeepfake}, Face2Face(F2F)~\cite{face2face}, FaceSwap(FS)~\cite{gitfaceswap} and NeuralTextures(NT)~\cite{nt}). The bold results are the best.}
\label{table:comparisonsingle_0.5}
\begin{tabular}{P{3.6cm} M{1.5cm} M{1.5cm} M{1.5cm} M{1.5cm}}
\hlinew{1.1pt}
Methods & DF~\cite{gitdeepfake} & F2F~\cite{face2face} & FS~\cite{gitfaceswap} & NT~\cite{nt} \\ 
\hline
Steg.Features~\cite{richmodel} & 67.00\%  & 48.00\%  & 49.00\% & 56.00\%  \\ 
LD-CNN~\cite{ff++17} & 75.00\% & 56.00\%  & 51.00\% & 62.00\%   \\
Constrained Conv~\cite{ff++10}& 87.00\% & 82.00\% & 74.00\% & 74.00\% \\
CustomPooling CNN~\cite{ff++51}& 80.00\% & 62.00\% & 59.00\% & 59.00\% \\
MesoNet~\cite{mesonet} & 90.00\% & 83.00\% & 83.00\% & 75.00\% \\
Xception~\cite{xception}& 94.27\% & 91.98\% & 93.03\% & 76.43\% \\ 
F$^{3}$-Net(Xception) & \textbf{96.81\%} & \textbf{94.01\%} & \textbf{95.85\%} & \textbf{79.36\%} \\ 
\hline \hline
Slowfast~\cite{slowfast} & 96.01\% & 92.47\% & 93.86\% & 80.07\% \\ 
F$^{3}$-Net(Slowfast) & \textbf{98.54\%} & \textbf{93.91\%} & \textbf{96.82\%} & \textbf{83.74\%} \\
\hlinew{1.1pt}
\end{tabular}
\end{table}

\subsection{Best Practice}

In this paper, we propose an innovative face forgery detection framework that can make use of frequency-aware forgery clues, named as F$^3$-Net.
%
%
Additionally, in this supplementary material, we discuss the the choice of hyper parameters in our framework.


\begin{table}[ht]
\centering
\caption{Comparison of the variants of LFS with different values of window sizes in SWDCT. The stride size is set as 2 and we train and test models in FF++ low quality (LQ) tasks. The bold results are the best.}
\label{table:windowsize}
\begin{tabular}{P{2cm} M{1.2cm} M{1.2cm} M{1.2cm} M{1.2cm} M{1.2cm}}
\hlinew{1.1pt}
Window size &  2 & 5 & 10 & 20 & 30  \\ \hline
AUC  & 0.836 & 0.883 & \textbf{0.889} & 0.876 & 0.853 \\ 
\hlinew{1.1pt}
\end{tabular}
\end{table}

\vspace{0.1cm}
\noindent\textbf{Window size \& stride size of LFS.} As described in Sec.3.2 in the main paper, Sliding Window DCT (SWDCT) is adopted in Local Frequency Statistics (LFS) branch to extract localized frequency responses. The window size and the stride size in sliding window of SWDCT need to be selected to get the best practice.
In this supplementary material, we further provide a discussion on different values of window sizes and stride sizes in SWDCT.

The comparison results of various window sizes of SWDCT among 2, 5, 10, 20 and 30 are listed in Tab. \ref{table:windowsize}. If LFS applies a small window size (\ie, 2), it cannot extract frequency statistics around edges and some other structures coped with high frequencies.
However, if the window size is too large (\ie, 20 and 30), the LFS is less sensitive to the local abnormal statistics thus there will also be a significant performance drop. We adopt the window size as 10 to get the best practice in our F$^3$-Net.


The stride size of SWDCT is also worth investigating as it is highly correlated with the detection performance and computational cost. We validate the value of stride among 10, 6, 4, 3, 2 and 1. We train and test models in FF++ low quality (LQ) tasks and the window size is set as 10.
%
By comparing the results listed in Tab. \ref{table:lfsstep}, we employ stride as 2 because it finds the best trade-off between effectiveness and efficiency.


\begin{table}[ht]
\centering
\caption{Comparison of the variants of LFS with different values of stride size in SWDCT. The window size is set as 10 and we train and test models in FF++ low quality (LQ) tasks. The time cost is averaged on 1000 runs. The bold results receive the best accuracy and AUC scores while the underline ones strike a balance between effectiveness and efficiency.}
\label{table:lfsstep}
\begin{tabular}{P{1.8cm}  M{1.2cm} M{1.2cm} M{1.2cm} M{1.2cm} M{1.2cm} M{1.2cm} M{1.2cm}}
\hlinew{1.1pt}
Stride size & 10 & 6 & 4 & 3 & 2 & 1 \\ \hline
AUC & 0.835 & 0.855 & 0.876 & 0.882 & \underline{0.889} & \textbf{0.891} \\ 
Time(s/iter) & 0.368 & 0.376 & 0.384 & 0.394 & \underline{0.402} & \textbf{0.832} \\ 
\hlinew{1.1pt}
\end{tabular}
\end{table}

\end{document}